\documentclass[12pt]{article}

\usepackage[margin=1in]{geometry}   % 1-inch margins
\usepackage{setspace}               % for double spacing
\usepackage{times}                  % Times font (common for journals)
\usepackage{authblk}                % for author affiliations
\usepackage[numbers]{natbib}        % numeric citations
\usepackage{graphicx}               % figures
\usepackage{amsmath, amssymb}       % math
\usepackage{xcolor}
\usepackage{url}
\usepackage{hyperref}

\DeclareMathOperator{\AUC}{AUC}
\newcommand{\MyDelta}{$|\Delta\AUC|$}

\title{Unlocking the power of partnership: How humans and machines can work together to improve face recognition}
\author[1]{P. Jonathon Phillips}
\author[2]{G\'eraldine Jeckeln}
\author[1]{Amy N. Yates}
\author[1]{Carina A. Hahn}
\author[1]{Peter C. Fontana}
\author[2]{Alice J. O'Toole}
\affil[1]{Technology Testing and Evaluation Division, National Institute of Standards and Technology, Gaithersburg, MD 20899, USA}
\affil[2]{School of Behavioral and Brain Sciences, The University of Texas at Dallas, Richardson TX, USA}

\date{}  % remove date

\begin{document}

\maketitle
\newpage

\begin{abstract}
Human review of consequential decisions by face recognition algorithms creates a collaborative human-machine system. We establish the circumstances under which combining human and machine face identification decisions improves accuracy. Using data from expert and non-expert face identifiers, we show that the benefits of human-human and human-machine collaborations increase as the difference in baseline accuracy between collaborators decreases. This rule holds across a wide range of baseline abilities, from novices to professional forensic face examiners. An important consequence of the rule is that people who are substantially less accurate than the machine, can actually {\it improve} decision accuracy when they collaborate with the machine.   In a group of individual people collaborating with a machine, ``intelligent human-machine fusion'' was implemented by selecting people with the potential to increase collaborative accuracy. Performance with intelligent human-machine fusion was more accurate than either the machine operating alone or fusing all humans with the machine. Eliminating the machine from consideration yielded less predictable results, with average performance at or below intelligent human-machine collaboration. However, intelligent human-machine fusion was consistently more effective at minimizing the impact of low-performing humans on accuracy. The results demonstrate a meaningful role for both humans and machines in assuring accurate face identification.
\end{abstract}
\begin{center}
\end{center}
\newpage
\section*{Introduction}

Errors in face identification can have serious consequences for individuals, including unfounded 
criminal accusations and denial of entry to a country. To minimize errors, facial examiners work in teams; at borders, human inspectors review the results of automated face recognition. 
Collaborative face identification decisions 
recall the old adage ``two heads are better than one.''  But, is this always true?  
And, what happens if one of the two ``heads'' is a computer?

Machines play an integral role in face identification systems, and so determining the best way to combine human and machine decisions is critical for ensuring accurate face recognition. Societal concern of using an AI system without meaningful human oversight, combined with the superiority of present-day automated face recognition over humans (in some cases), makes this a pressing concern  \cite [e.g.,][]{phillips2018PNAS}. It is well-documented that human ``supervision'' of a face recognition system can decrease its accuracy  \cite{carragher2024trust,mueller2024automated}. People occasionally override correct machine decisions, even when they know that they are generally less accurate than the machine \cite{carragher2024trust,mueller2024automated}.

Despite these issues, there is ample evidence  that face identification accuracy generally increases when human and machine judgments  are combined \cite{o2007fusing,phillips2018PNAS}. 
%There is also evidence that accuracy is increased when two humans collaborate in making face identification decisions
Face identification  accuracy also generally improves when two humans collaborate on decisions
\cite{Cavazos2023CollaborationTI,dowsett2015unfamiliar,Jeckeln2018,balsdon2018improving,phillips2018PNAS,white2013crowd}. Although the benefit of fusion for face identification replicates across studies, within each study, not every human-human or human-machine ``collaboration'' increases identification accuracy. In any given case, combining the decisions of individual performers (human-human, human-machine) can increase accuracy, decrease it, or leave it unchanged \cite{phillips2018PNAS,White:2015aa}.

The adage ``two heads are better than one'' conjures images of two people working together. However, it is possible to combine the judgments of the two people working independently \cite{koriat2012two}. This leads to two kinds of fusion.
%Two kinds of fusion are common for face recognition in applied settings.
In social fusion, two people actively engage with each other  to make decisions. 
In independent fusion, each collaborator makes a response independently and the responses are subsequently 
combined (fused). 
%  Independent fusion works best when individual identification responses are made using ratings  of the certainty that 
% two images show the same identity versus  
% different identities.
 Independent fusion works best when individual identification respond by rating on a scale. Rating-based responses implicitly capture 
the confidence of the observer on each trial  \cite{bahrami2010optimally} and
can be fused by simple averaging.
 Because machines respond on an analogous scale 
(computed similarity between the machine's representations of the  images) \cite[cf.,][for a review]{o2021face}, human  ratings support
transparent comparisons between human and machine face identification decisions.
Social collaboration and independent fusion
 produce comparable gains in accuracy 
 over the individuals 
operating alone \cite{Cavazos2023CollaborationTI,Jeckeln2018}. Because independent fusion has the advantage of being applicable to  both human-human and human-machine collaborations, 
we focus on independent fusion.

In this work, we begin with the finding that  face identification fusion generally yields benefits, but that these benefits are unpredictable on a case-by-case basis. We define the fusion benefit as the difference in accuracy between a fused pair (human-human or human-machine) and the more accurate individual. When examining specific pairings, one key factor in success is the difference in ability between the two collaborators.  
This has been studied in  visual perception tasks \cite{bahrami2010optimally,bang2014does,koriat2012two} and for medical diagnoses \cite{Kurvers2016}. 
For visual detection, when collaborators work together, fusion is beneficial only when the observers have nearly equal visual sensitivity \cite{bahrami2010optimally}. For human teams with very different visual sensitivities, ``two heads are actually worse than the better one.'' This result has been replicated for visual perception judgments made independently \cite{koriat2012two}. In  medical decision-making, fusing the independent judgments of doctors yields better performance than the best doctor in a group {\it only} when the diagnostic accuracy of doctors is  similar \cite{Kurvers2016}. This finding is based on doctors' diagnoses  of breast cancer (mammograms) and skin cancer (skin lesions), using data from more than 20,000 real-world diagnoses from 140 doctors.

The results of these previous studies converge 
 on a remarkably simple rubric for 
fusing  human decisions: Human judgments
 should be combined only when the baseline accuracy of the observers
  is similar. We term this the \emph{proximal accuracy rule (PAR)}.  The PAR has never been investigated for human-machine collaborations, nor has it been examined for human-human collaborations for face recognition.
  In previous human-machine collaboration studies for face recognition, human participants made the final decision and had
{\it a priori}
knowledge about the baseline accuracy of the machine \citep[cf.,][]{bartlett2024benchmarking,carragher2024trust,mueller2024automated}. 
  A natural consequence of the PAR, if it holds for human-machine collaborations, is that a person can potentially improve  collaborative  accuracy, regardless of whether they perform more or less accurately than the machine. 
  We refer to the range of accuracy differences between collaborators that can benefit overall accuracy as the  \emph{critical fusion zone}. When a machine performs more accurately than the person overseeing it, this critical fusion zone is of special importance for knowing whether/how to combine human and machine judgments \cite{carragher2023simulated,phillips2018PNAS}.

 In the first part of the study, we tested whether the PAR predicts success in human partnerships for
  face identification and whether it predicts success in human-machine partnerships. We use data from the human-machine fusions  to define a critical fusion zone for the task. 
In the second part of the study, we implemented PAR-guided fusion across a group of individuals, each working in collaboration with a machine. 
This  is a common practice  in security and law enforcement scenarios. We examined the extent to which
``intelligent fusion'', guided
by the PAR, can improve the performance of entire group of human-machine  collaborations. 
We compared this {\it intelligent 
human-machine  system} to a {\it generic human-machine  system}, in which the decisions of every individual were fused with a machine. 
In a third condition, we eliminated the machine from the system and examined a group of individuals formed into {\it optimal human partnerships}. 
For these human-only partnerships,
we optimized teams composed of human dyads, using graph-matching theory \citep{edmonds1965maximum, edmonds1965paths}.  The  performance of {\it individuals} operating alone served as a control condition against which these other fusion strategies were compared.
The findings point to a meaningful role for both humans and  machines in assuring accurate face identification. 

\section*{Methods}
\subsection*{Data Source}

Human-human fusion and human-machine fusion 
 were evaluated with  face-identity matching tasks. We used existing data from the Expertise in Facial Comparison Test (EFCT) \citep{White:2015aa} and the Facial Expertise Test (FET) \citep{phillips2018PNAS}.
Both tests included  human participants across 
a wide range of skill levels from untrained university students to highly trained professional forensic face examiners. 
Participants viewed pairs of images and rated their certainty that the images in the pair showed the same person versus different people. For the machine judgments,
identification ``certainty'' was measured as
the similarity between machine-generated  representations of the two  images in the pair. 
 We  present results from the EFCT in the main paper and results for the FET task  in the Supplementary Materials Section.
The FET analysis replicates the findings for human-machine collaboration.

\begin{figure}%[htbp]
\begin{center}
\includegraphics[width=2.3in]{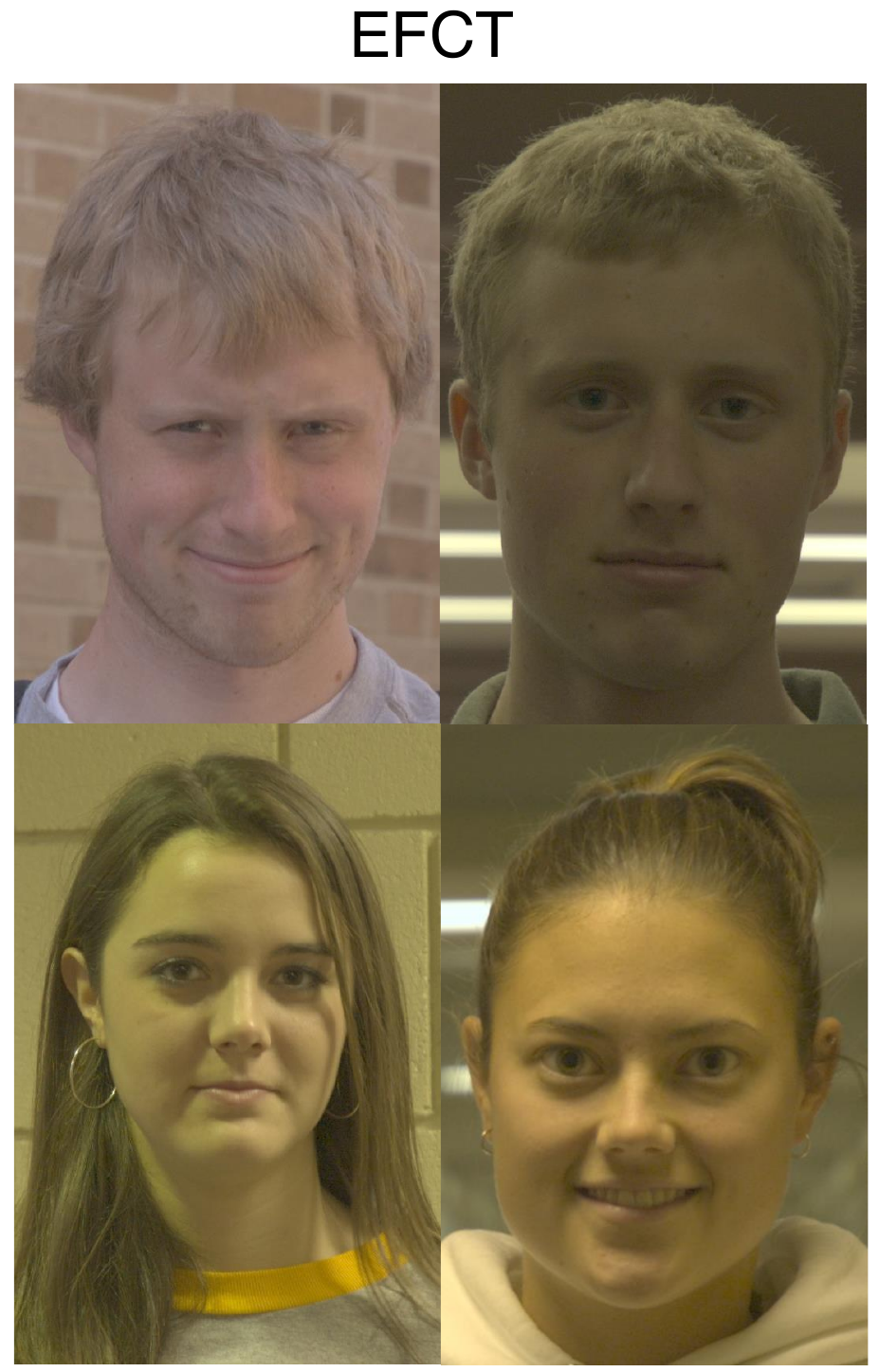}  

\caption{Example image-pairs from the EFCT. The top row shows same-identity face image pairs and the bottom row shows different-identity pairs (faces cropped from full image). }
\label{fig:F1_Example_EFCT_image-pairs}
\end{center}
\end{figure}

The EFCT includes 84 pairs of face images (42 same-identity and 42 different-identity pairs). Figure~\ref{fig:F1_Example_EFCT_image-pairs} shows example face image pairs.  We used human responses from the condition in which stimuli were presented upright for 30 seconds \cite{White:2015aa}.
Participants judged whether the image pairs showed the same identity or different identities,
 using a 5-point scale (1: sure they are the same person; 5: sure they are different people). The participant sample ($n=73$) included 27 forensic face examiners, 32 university students, and 14 participants attending a Facial Identification Scientific Working Group (FISWG) meeting \citep{White:2015aa}. The FISWG participants were selected as controls due to their interest in the topic and motivation to perform well. These participants were not trained in face identification.

For the machine data,  a deep convolutional neural network (DCNN) trained for face identification  \citep{parkhi2015deep} was  tested on the EFCT to obtain machine-generated similarity measures for each face-image pair. The machine was used to produce a face representation (i.e., embedding) for each image.

\subsection*{Fusion Benefit Metrics}
\label{subsec:fusion_benefit_metrics}

 We evaluated baseline human accuracy using the participants' independent responses for each image pair. Similarly, we evaluated the baseline machine accuracy using the  machine-generated similarity scores for each image pair.
 In both cases, face identification accuracy was measured as the area under the Receiver Operating Characteristic Curve (AUC). 
 An AUC of 0.50 indicates chance performance; 1.0 indicates perfect performance. 
 The human AUC was computed from the distributions of human face-matching responses for same-identity image pairs and different-identity image pairs. The machine AUC was computed from  
the distributions of machine-generated similarity scores for  same-  and different-identity pairs.

Fusion was achieved by combining the responses from two human observers (human-human fusion) or the  responses from a human and the machine (human-machine fusion). 
For human-human fusion, we investigated humans working together in teams of two, or dyads, in order to simulate how forensic facial examiners work in teams of two. Thus, human observers were assigned to
dyads, using methods that varied by the  analyses we report in different conditions of this study (e.g., all possible combinations, optimal pairing with graph matching). Next, for each dyad, we averaged the responses of the two people for each image pair in the face matching test. The fused accuracy  of the dyad ($\AUC_{fused}$) was computed from these averaged responses.

For human-machine fusion, human observers were paired with the machine to create dyads. Again the methods for pairing varied by condition (e.g., all possible pairs, intelligent fusion).  For each face-image pair ($i$), the  similarity between the machine-generated embeddings was computed, $s_i$, as the distance ($L_1$-norm) between the two embedding vectors.  
%Next, $s_i$ was scaled to a new value, $\widehat{s}_i$, to align it to the distribution of responses from all human participants,  as follows:
Next, $s_i$ was scaled to a new value, $\widehat{s}_i$, to align the shape of the model distribution to the range of the distribution of responses from all human participants,  as follows:

\[
\widehat{s}_i = \left( \frac{s_i - s_{\min}}{s_{\max} - s_{\min}} \right)(h_{\max} - h_{\min}) + h_{\min},
\]

\noindent

\noindent
where $s_{\min}$ and $s_{\max}$ denote the minimum and maximum of the original machine similarity scores, and $h_{\min}$ and $h_{\max}$ denote the minimum and maximum of human responses. For each human-machine dyad, on each face-image pair, we averaged the machine's scaled similarity score ($\widehat{s}_i$) and the human's response.

Absolute difference in baseline accuracy between the two participants (human-human or human-machine) in each dyad, \MyDelta, was computed as: 
\[
|\Delta\AUC| = |\AUC_{performer 1} - \AUC_{performer 2}|\
\]
Fusion benefit was measured by subtracting the accuracy of dyad's better performer ($AUC_{better}$) from the dyad's fused accuracy ($\AUC_{fused}$):
\[
\ \AUC_{fused} - \AUC_{better}\
\]

\subsection*{Optimal Human Dyads with Graph Matching}

Optimal human dyads were found using graph theory.  Note that
in human-machine fusion, we 
assumed that one machine was available to partner with
multiple humans.
In a completely human 
system (e.g., a group of forensic facial examiners), a single human can serve in only one dyad.
Therefore, the goal is to form human
dyads to yield the highest group performance, which we defined
as the average fused AUC for the selected dyads.   

From $N$ people,  we can make $N/2$ dyads (rounding down when $N$ is odd).  
To find these dyads, we turn to a branch of mathematics called graph theory. To solve for the optimal set of dyads, we formulated the problem as a weighted graph matching problem from graph theory \citep{edmonds1965maximum, edmonds1965paths}, see Supplementary Materials. As input to a weighted graph matching
formulation of this problem, we used fused AUC for all possible dyads.  

\begin{figure}
     \centering
         \includegraphics[width= 4.5in]{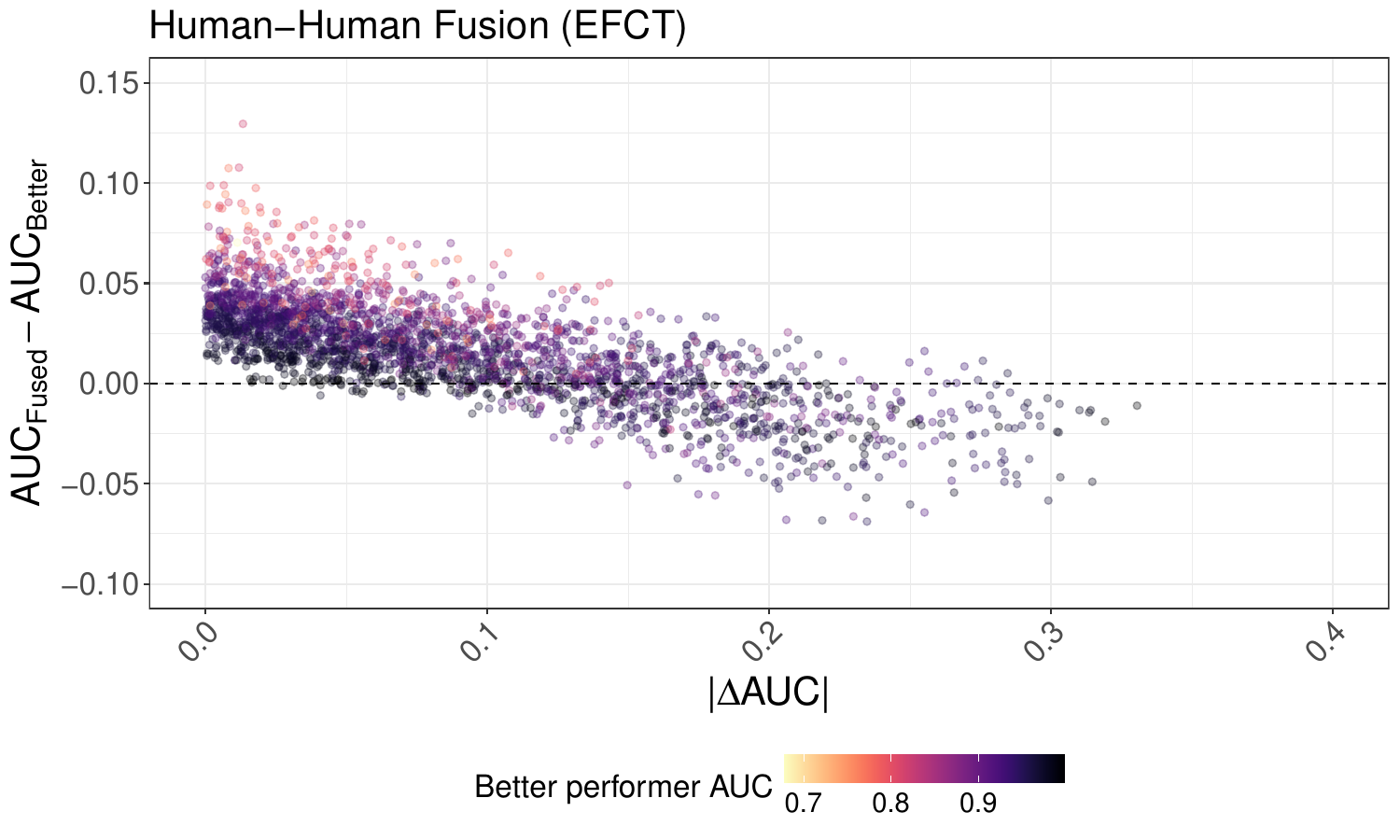}
       \\
        \includegraphics[width= 4.5in]{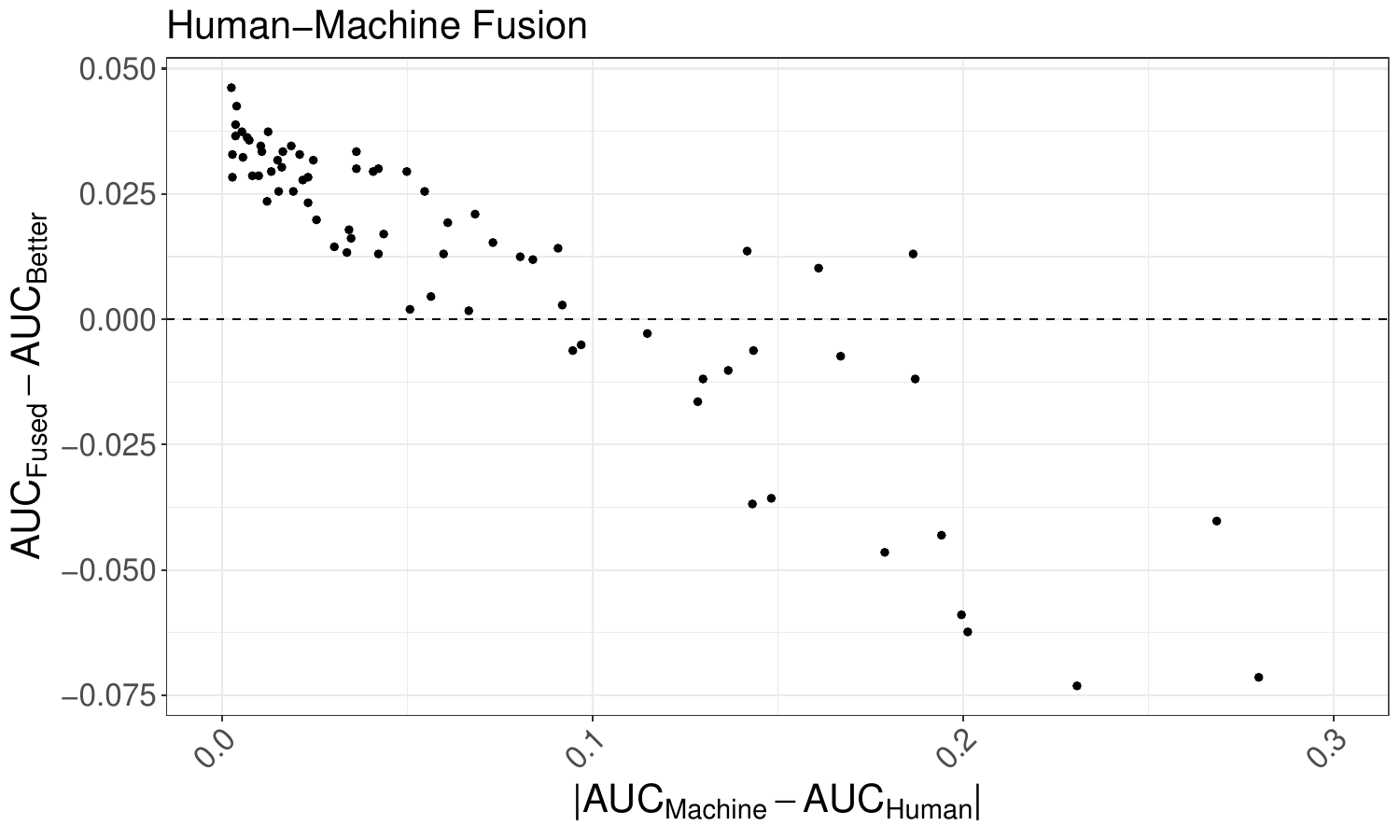}
        
         \caption{\textbf{Fusion benefits for human-human (top) and human-machine dyads  (bottom).} Fusion benefit (vertical axis) of each dyad (dot) plotted as a function of the absolute difference in accuracy (horizontal axis) between dyad members. Fusion benefit decreased with increased absolute difference in the baseline performance of the dyad members for both the human-human  (top panel) and human-machine (bottom panel) dyads.  %For dyads that above the horizontal lines, fusion is more beneficial than accepting the better performer's decision. 
         For dyads above the horizontal lines, the benefits of fusion surpass accepting the better performer's decision. For dyads below the horizontal lines, the advantages of accepting the better performer outweighs fusion.}
        \label{fig: predicting the benefit of fusion}
\end{figure}

\section*{Results}
\subsection*{Proximal Accuracy Rule for Face Identification}

First, we establish the PAR for face identification decisions made by human dyads. This  replicates and extends  similar results from other domains (e.g., medical decision-making \cite{Kurvers2016}). Second, we show that human-machine fusion success is also predicted by the PAR. This provides the first evidence that success in individual collaborations between a human and a machine can be predicted. 

\subsubsection*{Human-Human Partners} In this condition, we measured the fusion benefit for all possible pairs of humans.
The PAR rule predicted the success of  
human-human fusions for face identification judgments.
Figure~\ref{fig: predicting the benefit of fusion} (top panel) shows that
the benefit of human-human fusion  decreased as the difference in accuracy between the two performers  increased. More formally, a strong negative correlation was found between \MyDelta \ between participants and fusion benefit  ($r(2626) = -0.739$ , $p < 0.001$ , 95\% CI $[-0.755, -0.721]$). 
The figure also shows that the result holds 
across a wide range of baseline accuracy levels from poor performance to excellent performance. This is
indicated  by the color of the dots in Figure~\ref{fig: predicting the benefit of fusion} (top panel), which
code the baseline accuracy of the better performer in
each dyad. The figure makes clear that 
 with smaller differences in
baseline accuracy, fusion benefits are robust, even when the better performer in the dyad performs poorly.

\subsubsection*{Human-Machine Partnerships}
\label{sec: Predicting benefit of fusion, human and machine}

In this condition, we measured the fusion benefit for all humans paired with the machine. The PAR rule also predicted the success of fusion for  human-machine dyads. 
Figure~\ref{fig: predicting the benefit of fusion} (bottom panel) shows that
the benefit of human-machine fusion  decreased as the difference in accuracy between the human and machine   increased. A strong negative correlation was found between \MyDelta \ and fusion benefit $r(71) = -0.9034$, $p < 0.001$, 95\% CI $[-0.938, -0.850]$.

To summarize, for face identification fusion in both human-human and human-machine dyads, the smaller the difference in baseline accuracy between performers, the larger the fusion benefit. 
Given a large enough disparity in baseline performance between the two performers, fusion can degrade dyad performance to a level well below that of the more accurate participant.
The result also indicates 
that humans who perform less accurately than the machine can increase performance through collaboration.
Next, we consider how much worse a human can perform than a machine and still benefit collaborative accuracy.

\begin{figure}
     \centering
         \includegraphics[width= 4.5in]{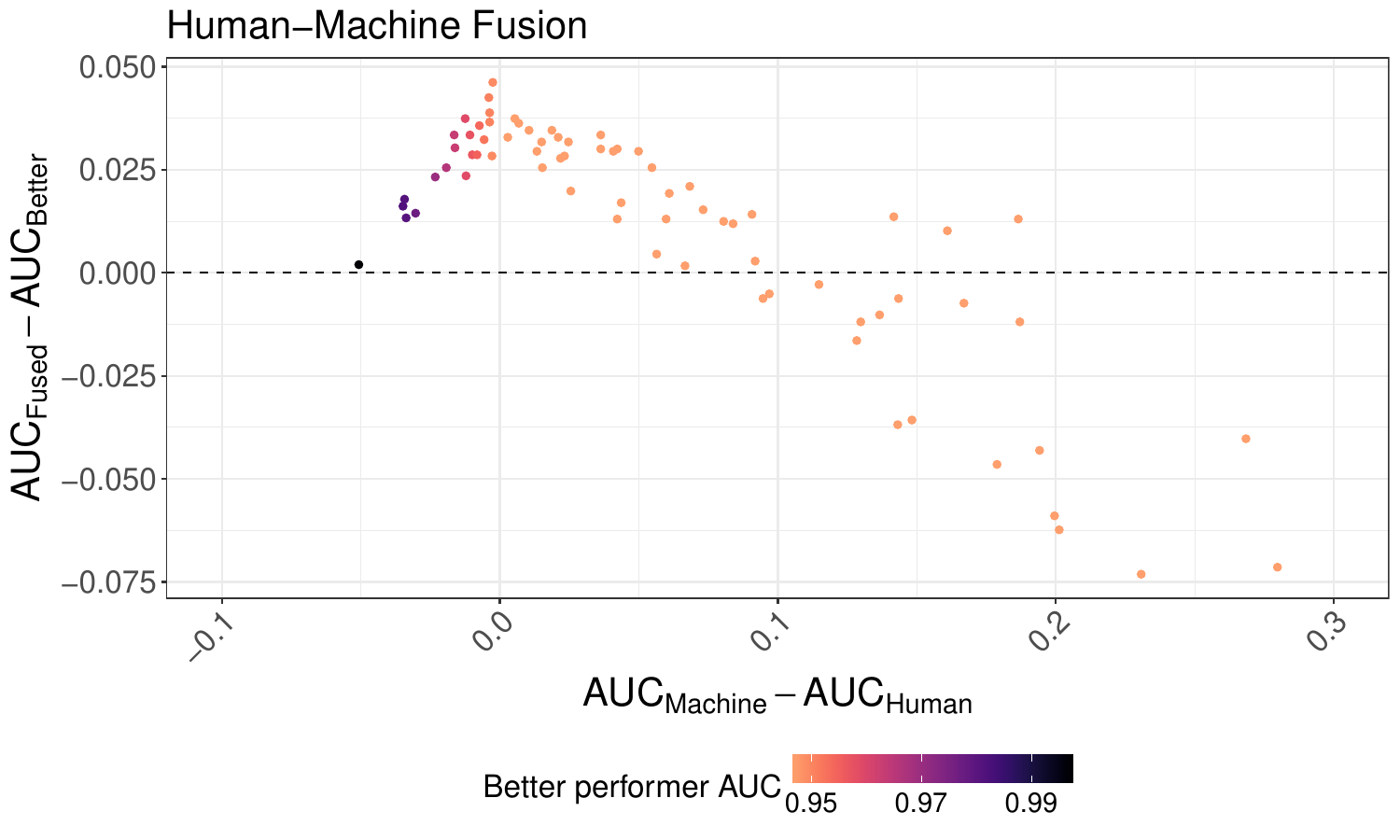}
         \caption{\textbf{Critical Fusion Zone.} The plot shows fusion benefit as a function of how much better/worse a human performed than the machine (machine at the zero point on the $x$-axis, machine and human performance are the same). Each point shows a human-machine fusion. The critical fusion zone is the range of $|\Delta\AUC| = |\AUC_{Machine} - \AUC_{Human}|$ values around the zero-point on the $x$-axis that yields fusion benefits (points above the dotted horizontal line).  Human-machine collaboration remains beneficial across a wide range of  human performance. Notably, this wide range includes many people who perform worse than the machine (points with positive values on both the $x$- and $y$-axes). The limit of the critical fusion  can be estimated near where the dots cross the horizontal dotted line  (fusion benefit = 0), and fusion benefit changes from positive to negative values. The simulations detailed in Section ``Humans, machines, both?'' determine this value precisely as 0.092 AUC units. Performance differences beyond this value yield fusion deficits.}
        \label{fig:critical_fusion_zone}
\end{figure}

\subsection*{Critical Fusion Zone}
 
Figure~\ref{fig:critical_fusion_zone}  shows fusion benefit as a function of how much more (or less) accurate the human is than the machine. The critical fusion zone encompasses the human participants who fuse beneficially with the machine (i.e., have a positive $y$-axis value in the figure).
% (We provide a precise value for the critical fusion zone in the next section.) 
As expected by the PAR, these human-machine dyads cluster around the point where machine and human accuracy are the same (0 on the $x$-axis). 
Included in the critical fusion zone are people who surpass machine accuracy (i.e., have a negative value on the $x$-axis) and  people who perform below the level of machine accuracy
(i.e., have a positive value on the $x$-axis).
Notably, performance benefits are found when fusing the machine with humans who perform substantially worse (0.092 AUC units) than the machine (see Figure \ref{fig:critical_fusion_zone}).    
Thus, for example, 
a human with a baseline accuracy of {$\AUC = 0.855$} can add to the accuracy of a machine that performs with accuracy of 
$\AUC = 0.947$.  
This surprisingly large
machine advantage is even larger for the FET test (see Supplementary Materials).
Crucially,  for a large majority of humans who perform beyond this critical difference, human-machine collaboration degrades performance relative to the machine operating alone.
Due to the high accuracy of the machine on this task, no human surpassed the machine by a margin that exceeded the critical fusion zone. As expected by the PAR, all humans who surpassed the  machine on this task yielded a fusion benefit when they collaborated with the machine.

Figure~\ref{fig:critical_fusion_zone} provides a first look at the extent of the critical fusion zone.  In the next section, we  examine the limits of the zone more precisely, using simulations based on a system of human-machine collaborators. 

\section*{Humans, Machines, Both?}
\label{sec: Benefits and Limits of Human-Machine Fusion}

A ``system'' is defined  as  
an entire group
of humans, working in any of the following configurations: a.) working with a machine, b.) working with a human partner, or c.) working alone.  
We compare these types of systems on the task of face identification. For human-machine systems, we considered intelligent human machine fusion and generic human-machine fusion.
 For intelligent human-machine fusion, humans were selectively paired with the machine, according to the PAR.
 For generic human-machine fusion, all humans were paired with the machine.  For the 
fully human system,   humans were paired  to optimize collaborative accuracy, using graph matching theory. In the control system, individuals worked independently. 

\section*{Intelligent Human-Machine Fusion System}
\label{section:intelligent_human_machine_fusion}

Beginning with intelligent human-machine fusion, the PAR tells us that we should fuse the decisions of a person and a machine  when their accuracy difference  falls within the critical fusion zone. Otherwise, we should defer to the more accurate person/machine in the pair.  
 {\it A priori} we know that optimal system performance will be reached when the fusion threshold is set to the critical zone. To determine the critical zone precisely, we examined the benefit/cost of selective fusion of humans and machines across a range of fusion thresholds.  Specifically, we calculated system-wide 
performance by setting the
\MyDelta \ threshold ($\lambda$) for fusion, while implementing a 
{\it selective fusion} rule of the
form:
    \[decision = \begin{cases} \mbox{fuse human and machine,} & \mbox{if } |\Delta\AUC| \le \lambda   \\ \mbox{accept machine decision,} & \mbox{otherwise} \end{cases} \]

Next,
we computed system-wide performance, as the average of AUCs
for the decisions made by the fused dyads and the decisions that defaulted to the machine.\footnote{Note: use of the machine decision as the default was dictated by the fact that humans outside of the fusion zone reduced the accuracy of the system.} %were always less accurate than the machine. 
The results
showed that peak system-level accuracy was achieved at a critical fusion value of $\lambda = 0.092$. 

\begin{figure}
     \centering
         \includegraphics[width= 7.0in]{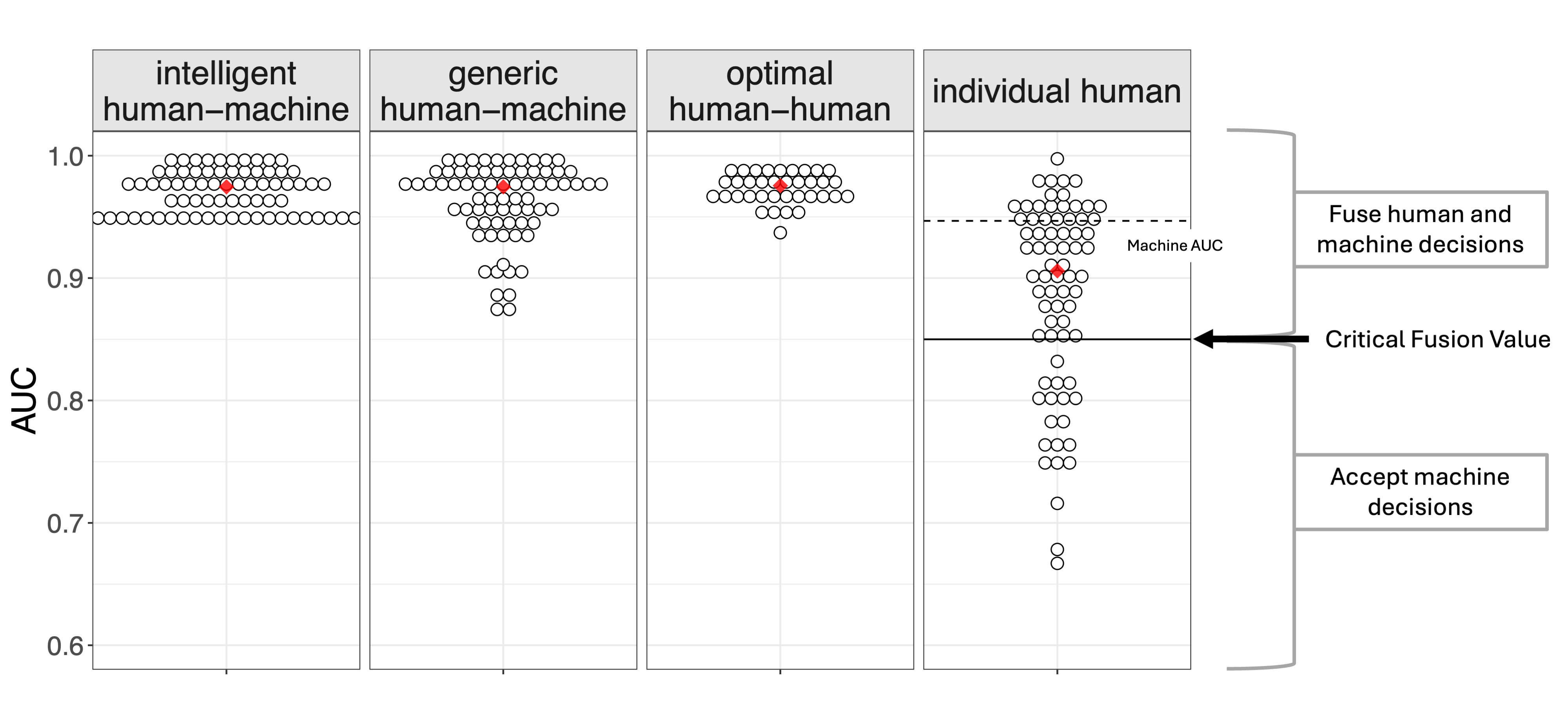}\\
        
        \caption{\textbf{Fusion Systems}. Performance for each of the systems. Intelligent human-machine fusion dots indicate fused dyads and dyads that defaulted to the machine. Generic human-machine fusion shows human-machine dyads. Optimal human-human fusion shows dyads identified through graph matching theory.  Individual human performance. The average performance for all fusions is comparable, but intelligent human-machine fusion results in the highest lower bound. The red diamonds represent the median for each system.}
        \label{fig:Analyses_all_EFCT}
\end{figure}

We implemented intelligent fusion across the human-computer dyads, fusing humans with the machine when the difference in their accuracy was less than or equal to the critical fusion value. 
Figure~\ref{fig:Analyses_all_EFCT} 
(first violin plot, leftmost) shows the results.  Each dot indicates the performance of either a human-machine fusion  ($|\Delta\AUC| \le \lambda$, critical fusion value) or the machine operating as a replacement for the human ($|\Delta\AUC| > \lambda$, critical fusion value). Average performance was AUC = 0.969. The lowest performance is bounded at the level of machine accuracy (M = 0.947).

\section*{Generic Human-Machine Fusion System}

What happens when humans are fused indiscriminately with the machine?
In the generic fusion condition, each human was fused with the machine.  Figure~\ref{fig:Analyses_all_EFCT} (second violin plot) shows the results. Average performance for generic human-machine performance (M = 0.962) was similar to that found for intelligent human-machine performance. However, the distribution of dyads extends well below that seen for intelligent human-machine fusion and the lower bound of performance is now inferior to the machine operating alone (minimum generic$_{AUC }$ = 0.874; minimum intelligent$_{AUC}$ = 0.947).
With generic fusion, 24.66\% of human-machine fusions fall below the performance of the machine.  Intelligent human-machine fusion was statistically superior to generic human-machine fusion, using a non-parametric statistical test due to the differences in the form of the distributions (see
Table~\ref{tab:EFCT_system_stats}, row 1).

\begin{table}[h]
\centering
\caption{\textbf{Comparing system-level accuracy across fusion methods.} Two non-parametric statistical tests are used for comparison: paired Wilcoxon Signed Rank test and independent Mann-Whitney, depending on whether the data are paired or not. HM stands for human-machine; HH stands for human-human.\\}
\begin{tabular}{l l l c c c}
\hline
\textbf{Condition 1} & \textbf{Condition 2} & \textbf{Test} & \textbf{Statistic} & \textbf{$p$} & \textbf{Bonferroni-adj. $p$} \\
\hline
Intelligent HM & Generic HM & Wilcoxon SR (paired) & $W = 27$   & $0.00222$ & $0.00667$ \\
Intelligent HM & Optimal HH & Mann--Whitney U      & $U = 1160$ & $0.335$   & $0.335$ \\
Intelligent HM & Individual & Wilcoxon SR (paired) & $W = 0$    & $< 0.001$  & $< 0.001$ \\
Generic HM     & Individual & Wilcoxon SR (paired) & $W = 0$    & $< 0.001$  & $< 0.001$ \\
Optimal HH     & Individual & Mann--Whitney U      & $U = 198$  & $< 0.001$  & $< 0.001$ \\
\hline
\end{tabular}
\label{tab:EFCT_system_stats}
\end{table}

\section*{Optimal human dyads from graph matching}

The optimal human pairs found by graph matching are plotted in  Figure~\ref{fig:Analyses_all_EFCT} (third violin plot). Performance for this condition is comparable to intelligent human-machine fusion. This is true for both the average AUC and the lower bound of the distribution, which falls slightly below that found for intelligent fusion (see
Table~\ref{tab:EFCT_system_stats}, row 2). Of note,
the FET  (See Supplementary Materials) shows a somewhat different result for this condition. Specifically, the optimal human dyad mean for the FET falls below its mean  for intelligent human-machine dyads. There is also a more extended distribution tail for the FET's optimal human pairings  than for its intelligent human-machine dyads. 
The treatment of human outliers in the optimal human dyad versus the intelligent human-machine conditions accounts for these differences.
Intelligent human-machine fusion 
allows for the machine to replace a human who performs  poorly.
Optimal human fusion, however, requires  all individuals to be included in a dyad. Therefore, individual differences among human performers makes
optimal human fusion  less  predictable than intelligent human-machine fusion.

In the Supplementary Materials,  we use permutations to show that the optimal dyads found through graph matching yield  more accurate performance than choosing human dyads at random.  

\section*{Individual human performance}
Individual performance is shown  in  Figure~\ref{fig:Analyses_all_EFCT} (fourth violin plot). Each dot shows the performance of an individual working alone.  
The figure clearly illustrates the power of fusion for improving average performance and for increasing the lower bound of performance by shrinking the tail of the distribution.
The plot shows the critical fusion value, as well as  the level of machine performance. The  median human performance (red dot) falls below machine performance, as does the performance of the majority of humans.  
The critical fusion value
can
be used to divide people into two categories: a.) fuse 
human and machine and b.) accept machine's decision.  Again,
it is clear from this figure that, within the limits set by the PAR, people who perform substantially
worse than the machine can contribute beneficially to system accuracy.

More formally,
nonparametric tests showed that intelligent and generic human-machine fusion, as well as optimal human-human fusion, were all statistically superior to individual humans working alone
(see Table~\ref{tab:EFCT_system_stats}).

\section*{Discussion}

  In an era where humans partner with AI
  to perform important and consequential tasks, it can be difficult to know what to do when a human and  machine disagree.  
In cases of conflict between two decisions, 
 it may seem intuitively reasonable to simply choose 
the better system (human or
machine). Or, if human review is a priority, it may seem best to  combine human and machine decisions. In both cases, this would be a mistake. If accuracy is the goal,
collaborative decision-making  offers a better alternative, when it is applied 
intelligently.

In this study, we show that intelligent fusion for face identification follows the proximal accuracy rule: decisions should be combined only when the baseline accuracies of the human and machine are similar.   Fusion benefits
increased linearly as the difference in baseline accuracy between the human and machine decreased. 
The  most important implication of these findings is that, within limits,  people who perform worse than a machine can  still 
contribute to the accuracy of its face identification decisions. 
The critical fusion zone defines the ``within-limits'' part of the human contribution to   fusion. It tells us how much worse a human can be than a machine and still improve decision accuracy.  The critical zones we found indicate that 
input from a human judge can be beneficial up to a surprisingly large machine advantage. 
% As such, indiscriminately
% discarding input from individuals with less ability than a machine would produce less accurate face identification overall.
Because
face identification technology is   now competitive with humans, this  finding has pragmatic value.
In head-to-head comparisons between humans and machines, machines are often at or near the top of the human performance distributions \cite{phillips2018PNAS,o2021face, jeckeln2024designing}---even for highly challenging tasks such as discriminating
the faces of identical twins \cite{parde2023twin}.  
In applied scenarios, 
many people, even face recognition experts, can be expected to perform less accurately than a machine.  The critical fusion zone indicates that lower performance by a person than a machine does not automatically disqualify the person's input. Instead,
PAR-guided fusion in these scenarios can  increase overall accuracy and serve the important societal benefit of keeping a human involved in the process. 

Collaborations between humans mirrored the effects  found for human-machine fusion. Fusion success was predicted by the PAR and there was a linear increase in fusion benefit as the difference in baseline abilities between observers decreased.  Fusion benefits were successfully predicted for
participants ranging from university students (no experience/poor ability) to professional forensic face examiners (extensive professional training/high ability).
The finding  that
the PAR applied equally to people with varying levels of ability suggests that it has the potential to improve accuracy in a wide range of applied scenarios, including those in which novice, rather than expert, face identifiers are  employed.  These results complement and extend  lessons learned from 
the  decision-making literature about when to
  combine human decisions 
\cite{bahrami2010optimally,koriat2012two,Kurvers2016}.    

In real world scenarios, collaborative benefits are best measured
by examining a group of humans and/or human-machine collaborators working in a system. In investigating these systems, we found that fusing every human with a machine (generic fusion) resulted in a large boost in accuracy over individuals working alone. This replicates previous findings indicating the general value of fusion
\cite{o2007fusing,phillips2018PNAS}. 
Additional performance gains were achieved with
intelligent PAR-guided fusion. Specifically,
intelligent fusion dramatically shortened the tail of the performance distribution and increased its lower bound. 
This tail-shrinkage is a natural 
consequence of the fusion decision strategy, which lets the machine act as a floor for the distribution of performance.
In applied face identification scenarios of consequence, system-wide performance  
is only as good as its weakest link. The ability of intelligent human-machine fusion to
shrink the tail of the individual performance distribution, therefore, has value
 in forensic applications. Studies with both
face experts and novices commonly show a broad range of individual differences in accuracy
\cite{davis2016investigating,norell2015effect,towler2017evaluating,phillips2018PNAS,White:2015aa,white2021understanding,wilmer2017individual}. Even novices can occasionally surpass  the performance of 
 experts  \cite{phillips2018PNAS,White:2015aa}. 
With intelligent fusion, the high performance of machines, relative to most humans, provides some 
level of protection against poor performance by human outliers.

Eliminating the machine from the system, and instead forming optimal human partnerships, had benefits that were less predictable. For the EFCT, optimal partnerships proved as effective as intelligent-human machine fusion. 
For the FET, intelligent human-machine fusion proved better than optimal human partners. Intelligent fusion of the FET produced  higher average performance than optimal human partners and shortened the tail of the distribution. The main conclusion from optimal human partnering is that under certain conditions, a fully human system can perform as well as an intelligent human-machine system. Under other conditions, intelligent human-machine fusion is the better choice.
One requirement for a comparable fully human system is that 
people be paired optimally. Random pairing of people yielded far worse fusion outcomes than optimal pairing (See Supplementary Material).  We used graph matching  to  optimize partnerships across system-wide human resources (i.e., people). As applied here, weighted graph matching was achieved by taking into account dyad fusion benefits across a group of human judges. A second requirement for a comparable  fully human system concerns the number and range of 
human outliers who perform poorly. 
Optimal human pairing includes all human judges and so poor performers may  limit fusion benefits.  

In the broader context, the fused systems we tested (intelligent human-machine fusion, generic human-machine fusion, and optimal human pairing) were all substantially more accurate  than humans working alone.
All showed higher  average performance and shorter distribution tails, thereby minimizing the impact of poor performers on system level performance. As such,  the practice of incorporating collaboration into face identification in applied scenarios can be a powerful strategy for increasing accuracy, even when it is not optimized. At the level of individual decisions, however, a strong case can be made for taking into consideration the relative abilities of the judges (human or machine). In consequential decisions in judicial and forensic settings, accuracy in individual cases is critically important. Thus, although it is valuable to know that fusion is an effective strategy in the general case, predicting accuracy on a case-by-case basis may protect individuals from harm and boost performance as a whole.  

% We found that the Proximal Accuracy Rule, which predicts fusion benefits in other diverse human decision tasks  \cite{bahrami2010optimally,koriat2012two,Kurvers2016}, 
% generalizes seamlessly to face identification. This was true for human-human collaborations and for human-machine collaborations.  
% The applicability of the PAR  across multiple diverse 
% tasks suggests that effective human oversight of  
% machine-generated decisions can come from
% leveraging knowledge about
% the baseline abilities of the individuals and machines 
% to-be-fused. 

% Combining judgments amounts simply to summing appropriately scaled decision certainties. The effectiveness and simplicity of a summation rule for fusion is consistent with longstanding theory in combining  the decisions of pattern classifier algorithms from different modalities (e.g., face, fingerprint, iris) \cite{KittlerCombiningClassifierTPAMI1998}.

In closing,
it is perhaps worth pausing to state the obvious;
the most accurate face identification will come from employing people and/or machines with the highest levels of skill. In judicial settings, for example, it is possible to select individuals with the highest level of skill and to assure adequate training. Professional forensic face examiners, reviewers, and super-recognizers surpass novices on face identification tests. Even in this best case scenario, there is a wide range of 
individual performance
 \cite[e.g.,][]{phillips2018PNAS, White:2015aa}.
In other consequential cases, for example, passport examiners at border crossings, the applicability of the PAR
across  skill levels makes it possible to improve face identification.

Returning to the original question of whether two heads are better than one for face identification, we conclude that this depends on the difference in baseline ability of the two performers. This is true even when one of the ``heads'' is a computer.  As machines contribute to socially consequential tasks, methods for combining human and machine decisions need to be based on theories that are valid for both humans and machines.  
% The PAR gave us the critical fusion zone, which shows that combining humans and machines is valuable even when human ability is significantly less than that of machines.

\section{Data Availability}
De-identified data and analysis code used in this study  can be accessed on the Open Science Framework (see link below).
%For the FET, de-identified data is available as supplemental material in \cite{phillips2018PNAS}. Student and algorithm data can be found in Datasets S1 and S2. Examiner, reviewer, superrecognizer data can be obtained by signing a data transfer agreement with the NIST.

 \url{https://osf.io/yzh42/overview?view_only=863f8e80475946d08cdbee8f4b74af9e}.

\bibliographystyle{unsrtnat}
\bibliography{main}  
\newpage

\section*{Supplementary Material}

% \begin{figure}%[htbp]
% \begin{center}
% \includegraphics[width=2.3in]{Figs/FET/Fig1_FET_face_example.pdf} 
% \caption{Example image-pairs from the FET The top row shows same-identity face image pairs and the bottom row shows different-identity pairs (faces cropped from full image). }
% \label{fig:Example_FET_image-pairs}
% \end{center}
% \end{figure}

\subsection*{SM1. Replication using the Facial
Expertise Test}

All analyses were conducted as described for the EFCT.  For human-machine fusion, human participants partnered with A2017b  \cite{ranjan2017l2}.

\subsubsection*{Facial
Expertise Test (FET)} 

The FET included 20 challenging face-image pairs (12 same-identity and 8 different-identity pairs) \cite{phillips2018PNAS}. %(see example in Figure~\ref{fig:Example_FET_image-pairs})
Again, the task was to determine whether the two images portray the same or different identities. For the human data, participants responded using a 7-point scale ($-3$: high confidence that the pair showed different people; +3: high confidence that the pair showed the same person). The human participants ($n=184$) included 57 forensic facial examiners, 30 facial reviewers, 13 super-recognizers, 53 fingerprint examiners, and 31 students (controls) \citep{phillips2018PNAS}. Fingerprint examiners were included as controls who were forensically trained on fingerprint identification, but not on  face identification. For the machine input, we used the  most accurate of the four DCNNs tested in previous work \cite{phillips2018PNAS}, 
A2017b \cite{ranjan2017l2}.  Human-machine fusion was performed as described for the EFCT.

\begin{figure}
     \centering
         \includegraphics[width= 4.5in]{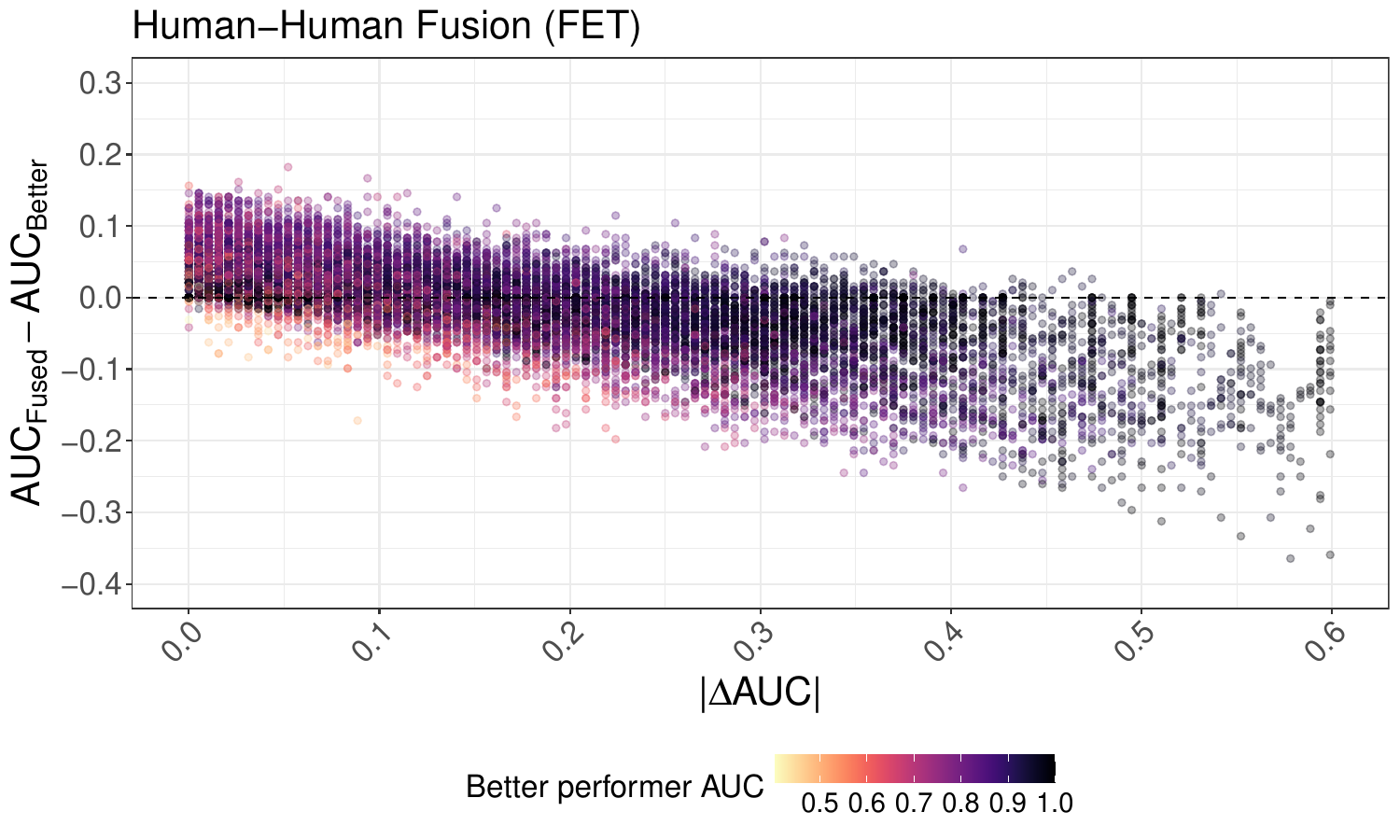}
       \\
        \includegraphics[width= 4.5in]{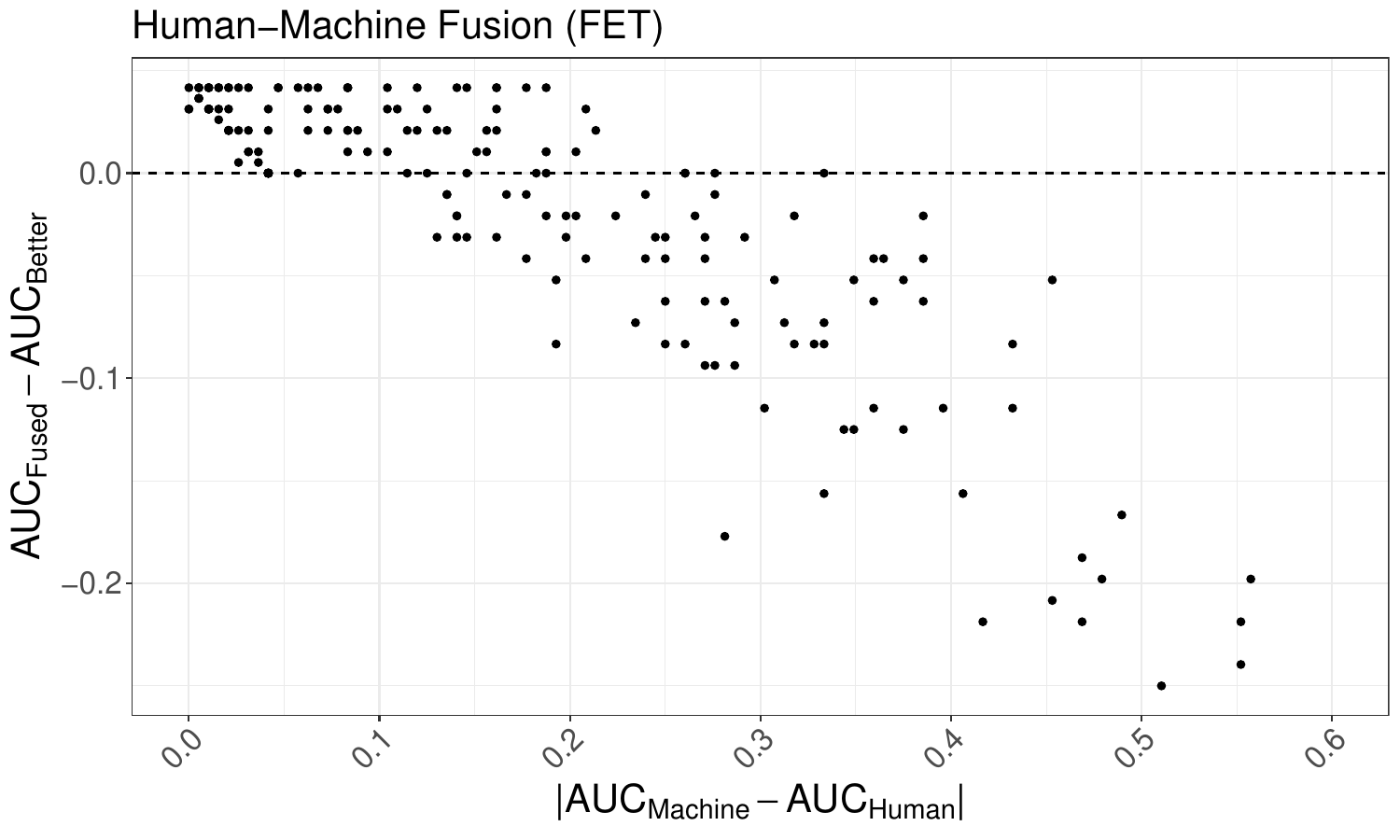}
        
         \caption{\textbf{Fusion benefits for human-human (top) and human-machine dyads  (bottom).} Fusion benefit of each dyad (dot) plotted as a function of the absolute difference in accuracy between dyad members. Fusion benefit decreased with increased difference in the baseline performance of the dyad members for both the human-human  (top panel) and human-machine (bottom panel) dyads.  For dyads that above the horizontal lines, fusion is more beneficial than accepting the better performer's decision.}
        \label{fig: FET predicting the benefit of fusion}
\end{figure}

\begin{figure}
     \centering
         \includegraphics[width= 4.5in]{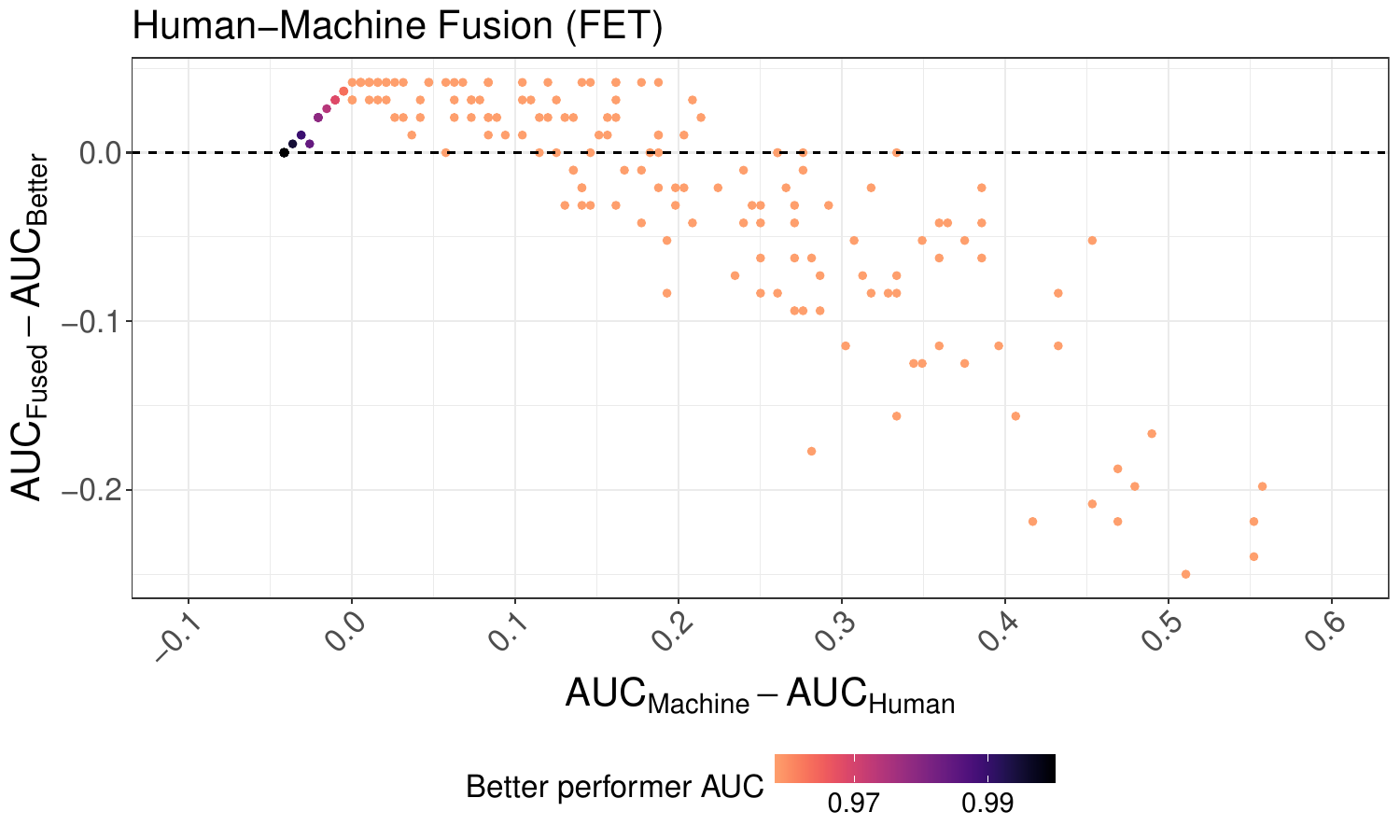}
         \caption{\textbf{Critical Fusion Zone.} %The plot shows fusion benefit as a function of how much better/worse a human performed than the machine (machine performance is at the zero point on the $x$-axis). Each point shows a human-machine fusion. The critical fusion zone is the range of \MyDelta \ values around machine performance that yields fusion benefits (points above the dotted horizontal line).  Human-machine collaboration remains beneficial across a wide range of  human performance. Notably, this wide range includes many   people who perform worse than the machine (points with positive values on both the $x$- and $y$-axes). The critical fusion value for people who perform worse than the machine can be estimated at 0.10 from this plot. Beyond this value most human-machine collaborations yield fusion deficits.
         }
        \label{fig:fet critical_fusion_zone}
\end{figure}

\subsubsection*{Results: Proximal Accuracy Rule for Face Identification}

The PAR rule predicted the success for both human-human and human-machine fusions for face identification judgments. As shown in Figure~\ref{fig: FET predicting the benefit of fusion} (top panel), 
results showed that the benefit of pairing two human observers declined as the difference in their accuracy increased ($r(16834) = -0.703$, $p < 0.001$,  95\% CI $[-0.711, -0.695]$). Consistent with this pattern, the benefit of pairing a human observer with a machine declined with greater disparities between human and machine performance
($r(182) = -0.853$, $p < 0.001$, 95\% CI $[-0.888, -0.808]$) (see Figure~\ref{fig: FET predicting the benefit of fusion}, bottom panel).

\subsubsection*{Descriptive Results: System-Level Fusion and Individual Performance}

Applying the selective fusion rule (see Section Intelligent Human–Machine Fusion System), peak system‑level accuracy was achieved at a critical fusion value of $\lambda$ = 0.188. We used this threshold to guide intelligent human–machine fusion. Performance under this condition was higher on average (intelligent human–machine fusion: \emph{Med} = 0.969, \emph{M} = 0.975, \emph{SD} = 0.021) than that observed for human observers working alone (\emph{Med} = 0.815, \emph{M} = 0.796, \emph{SD} = 0.153) and for the machine operating alone (0.9583). Implementing the critical fusion value showed that human observers with baseline accuracy corresponding to a minimum of AUC = 0.7703 can still contribute positively to system performance. Intelligent human-machine fusion also showed higher average performance than generic human–machine fusion (\emph{Med} = 0.969, \emph{M} = 0.944, \emph{SD} = 0.069).

\subsubsection*{Statistical Comparisons Across System Configurations}

Nonparametric tests were used to compare system-level accuracy across fusion methods. See Table SM1 for results. Wilcoxon signed-rank tests were applied for paired comparisons, and Mann--Whitney U tests for unpaired comparisons. All reported $p$ values are Bonferroni corrected. Overall, intelligent human-machine fusion outperformed generic fusion, individual performance, and the optimal human-human system. Generic fusion also exceeded individual performance.
\begin{table}[h]
\centering
\caption{\textbf{Compare system-level accuracy across fusion methods.} HM stands for human-machine; HH stands for human-human.\\}
\begin{tabular}{l l l c c c}
\hline
\textbf{Condition 1} & \textbf{Condition 2} & \textbf{Test} & \textbf{Statistic} & \textbf{$p$} & \textbf{Bonferroni-adj. $p$} \\
\hline
Intelligent HM & Generic HM & Wilcoxon SR (paired) & $W = 25.5$ & $< 0.001$ & $< 0.001$ \\
Intelligent HM & Optimal HH & Mann--Whitney U      & $U = 6860$ & $0.00904$ & $0.00904$ \\
Intelligent HM & Individual & Wilcoxon SR (paired) & $W = 0$    & $< 0.001$  & $< 0.001$ \\
Generic HM     & Individual & Wilcoxon SR (paired) & $W = 0$    & $< 0.001$  & $< 0.001$ \\
Optimal HH     & Individual & Mann--Whitney U      & $U = 2610$ & $< 0.001$  & $< 0.001$ \\
\hline
\end{tabular}
\label{tab:FET_system_stats}
\end{table}

\begin{figure}
     \centering
         \includegraphics[width= 7.0in]{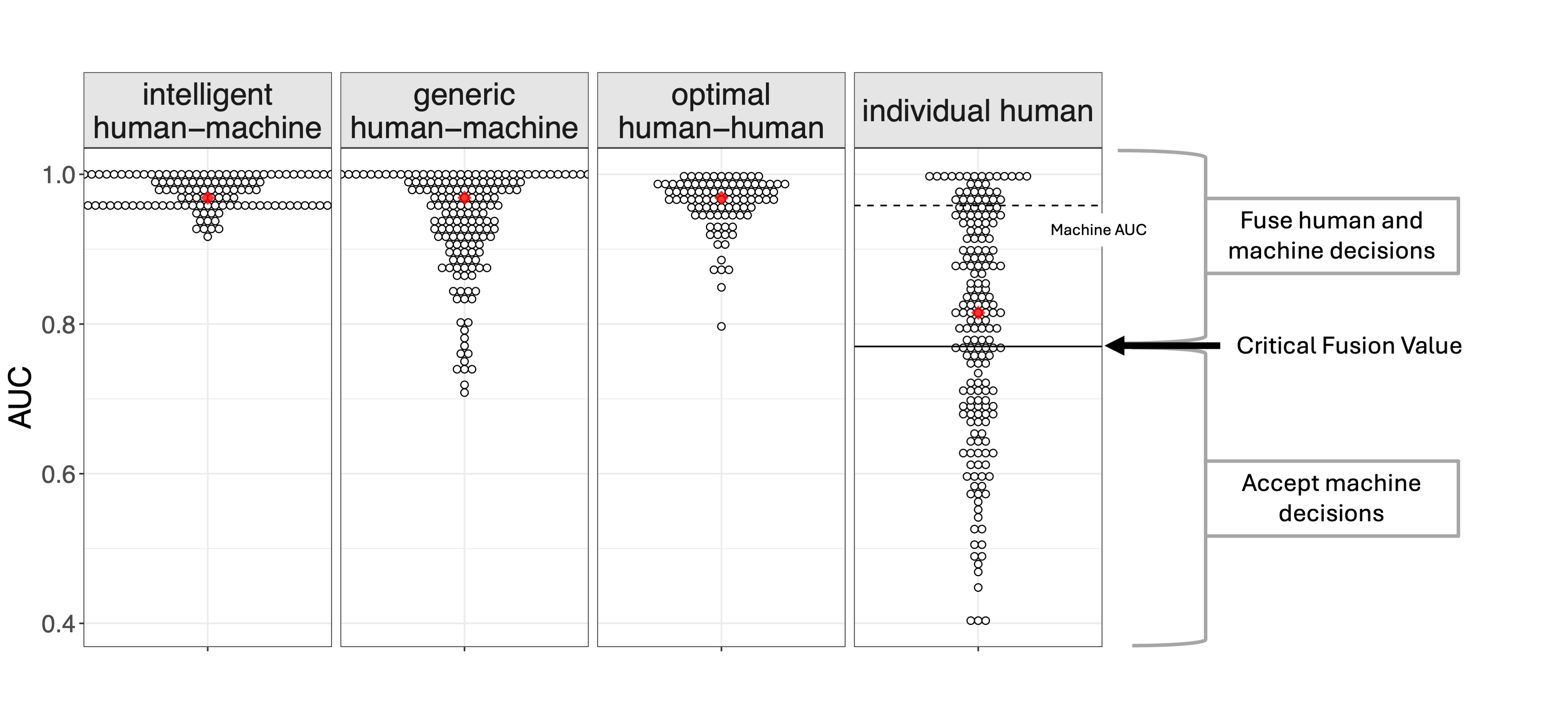}\\
        
        \caption{\textbf{Fusion Systems}. Performance for each of the systems. Intelligent human-machine fusion dots indicate fused dyads and dyads that defaulted to the machine. Generic human-machine fusion shows human-machine dyads. Optimal human-human fusion shows dyads identified through graph matching theory.  Individual human performance.}
        \label{fig:Analyses_all_FET}
\end{figure}

\subsection*{SM2. Additional Methods and Results (EFCT and FET)}

\paragraph{Graph Matching} 

To assess whether optimal human partnering is substantially better than randomly selecting partners, we generated 100 systems consisting of random dyads. We enforced the rule that a person can only be in one dyad. The mean was computed for each of the random system. For the EFCT and FET, there was no overlap between the optimal system mean and the accuracy distribution of the means for the random systems. 
For the EFCT, the optimal system achieved an accuracy of $0.973$.
Random system accuracy (Mean = $0.9463$; $sd = 0.0026$) differed from the optimal system accuracy by 
$10.40$ standard deviations.
For the FET, optimal human–human system obtained via graph matching achieved an accuracy of $0.961$.
Random system accuracy 
(Mean $=$
$0.874$; %0.8738615
sd $=$ 
$0.0040$)  %0.003959789
differed from the optimal system accuracy by $22.0615$ % 22.0615
standard deviations. The optimal system was far superior to randomly partnering people.

\end{document}